\documentclass{article}

\usepackage{amsmath,amssymb,amsthm}
\usepackage{booktabs} 
\usepackage[british]{babel}
\usepackage{comment}
\usepackage[T1]{fontenc}
\usepackage[utf8]{inputenc}
\usepackage{mathtools}
\usepackage{microtype}
\usepackage{graphicx}
\usepackage{microtype}      
\usepackage{nicefrac}
\usepackage{physics}
\usepackage{subfigure}
\usepackage{wrapfig}

\usepackage[backref=page]{hyperref}

\usepackage[accepted]{icml2023}

\usepackage[capitalize,noabbrev]{cleveref}

\definecolor{C0}{HTML}{1f77b4}
\definecolor{C1}{HTML}{ff7f0e}
\definecolor{C2}{HTML}{2ca02c}
\definecolor{C3}{HTML}{d62728}
\definecolor{C4}{HTML}{9467bd}
\definecolor{C5}{HTML}{8c564b}
\definecolor{C6}{HTML}{e377c2}
\definecolor{C7}{HTML}{7f7f7f}
\definecolor{C8}{HTML}{bcbd22}
\definecolor{C9}{HTML}{17becf}
\definecolor{gaussian}{HTML}{47908C}

\newtheorem{theorem}{Theorem}

\newtheorem{conjecture}[theorem]{Conjecture}

\newcommand{\ka}{\kappa}

\newcommand{\EE}{\mathbb{E}\,}
\newcommand{\EEp}{\underset{+}{\mathbb{E}}\,}
\newcommand{\EEm}{\underset{-}{\mathbb{E}}\,}
\newcommand{\reals}{\mathbb{R}}

\icmltitlerunning{Neural networks trained with SGD learn distributions of increasing complexity}

\begin{document}

\twocolumn[
\icmltitle{Neural networks trained with SGD learn distributions of increasing complexity}



\icmlsetsymbol{equal}{*}

\begin{icmlauthorlist}
\icmlauthor{Maria Refinetti}{ens,epfl}
\icmlauthor{Alessandro Ingrosso}{ictp}
\icmlauthor{Sebastian Goldt}{sissa}
\end{icmlauthorlist}

\icmlaffiliation{ens}{Laboratoire de Physique de l’Ecole Normale Supérieure, Université
  PSL, CNRS, Sorbonne Université, Université Paris-Diderot, Sorbonne Paris
  Cité, Paris, France}
\icmlaffiliation{epfl}{IdePHICS laboratory, \'Ecole F\'ed\'erale Polytechnique de Lausanne (EPFL), Switzerland}
\icmlaffiliation{ictp}{The Abdus Salam International Centre for Theoretical Physics (ICTP), Trieste, Italy}
\icmlaffiliation{sissa}{International School of Advanced Studies (SISSA), Trieste, Italy}

\icmlcorrespondingauthor{Sebastian Goldt}{sgoldt@sissa.it}

\icmlkeywords{Machine Learning, ICML}

\vskip 0.3in
]



\printAffiliationsAndNotice{}  

\begin{abstract}
The uncanny ability of over-parameterised neural networks to generalise well
has been explained using various ``simplicity biases''.
These theories postulate that neural networks avoid overfitting by first fitting
simple, linear classifiers before learning more complex, non-linear
functions.
Meanwhile, data structure is also recognised as a key ingredient for good
generalisation, yet its  role in simplicity biases is not yet understood.
Here, we show that neural networks trained using stochastic gradient descent
initially classify their inputs using lower-order input statistics, like mean
and covariance, and exploit higher-order statistics only later during training.
We first demonstrate this \emph{distributional simplicity bias} (DSB) in a
solvable model of a single neuron trained on synthetic data.
We then demonstrate DSB empirically in a range of deep convolutional networks
and visual transformers trained on CIFAR10, and show that it even holds in
networks pre-trained on ImageNet.
We discuss the relation of DSB to other simplicity biases and consider its
implications for the principle of Gaussian universality in learning.


\end{abstract}


\section{Introduction}%
\label{sec:introduction}

The success of neural networks on supervised classification tasks, and in
particular their ability to simultaneously fit their training data and
generalise well, have been explained using various ``simplicity
biases''~\citep{saad1995a, farnia2018spectral, valle-perez2018deep,
  kalimeris2019sgd, rahaman2019spectral}.  These theories postulate that neural
networks trained with stochastic gradient descent (SGD) learn ``simple''
functions first, and increase their complexity only as far as this is required
to fit the data. This bias towards simple functions would prevent the network
from overfitting, which is all the more remarkable given that the training loss
of modern neural networks has global minima with high generalisation
error~\citep{liu2020bad}.

Simplicity biases take various forms. Linear neural networks with small initial
weights learn the most relevant directions of their target function
first~\citep{le1991eigenvalues, krogh1992generalization, saxe2014exact,
  saxe2019mathematical, advani2020highdimensional}. Non-linear neural networks
learn increasingly complex functions during training, going from simple linear
functions to more complex, non-linear functions. This behaviour has been
demonstrated theoretically in two-layer
networks~\citep{schwarze1992generalization, saad1995a, engel2001statistical,
  Mei2018} and was confirmed experimentally in convolutional networks trained on
CIFAR10 by \citet{kalimeris2019sgd}. There is also evidence for a
\emph{spectral} simplicity bias, whereby lower frequencies of a target function,
or the top eigenfunctions of the neural tangent kernel, are learnt first during
training~\citep{xu2018understanding, farnia2018spectral, rahaman2019spectral,
  jin2020implicit, bowman2022implicit, yang2022overcoming}.

Meanwhile, data structure -- for example the low intrinsic dimension of
images~\citep{pope2021intrinsic} -- has been recognised as a key factor enabling
good generalisation in neural networks both empirically and
theoretically~\citep{mossel2016deep, bach2017breaking, goldt2020modelling,
  spigler2020asymptotic, ghorbani2020neural, pope2021intrinsic}. However, its
role in simplicity biases is not yet understood. 

Here, we propose a \emph{distributional simplicity bias} that shifts the focus
from characterising the function that the network learns, to identifying the
features of the training data that influence the network. We conjecture that any
parametric model, trained on a classification task using SGD, is initially
biased towards exploiting lower-order input statistics to classify its
inputs. As training progresses, the network then takes increasingly higher-order
statistics of the inputs into account.

\begin{figure*}[t]
  \centering
  \includegraphics[width=.95\linewidth]{{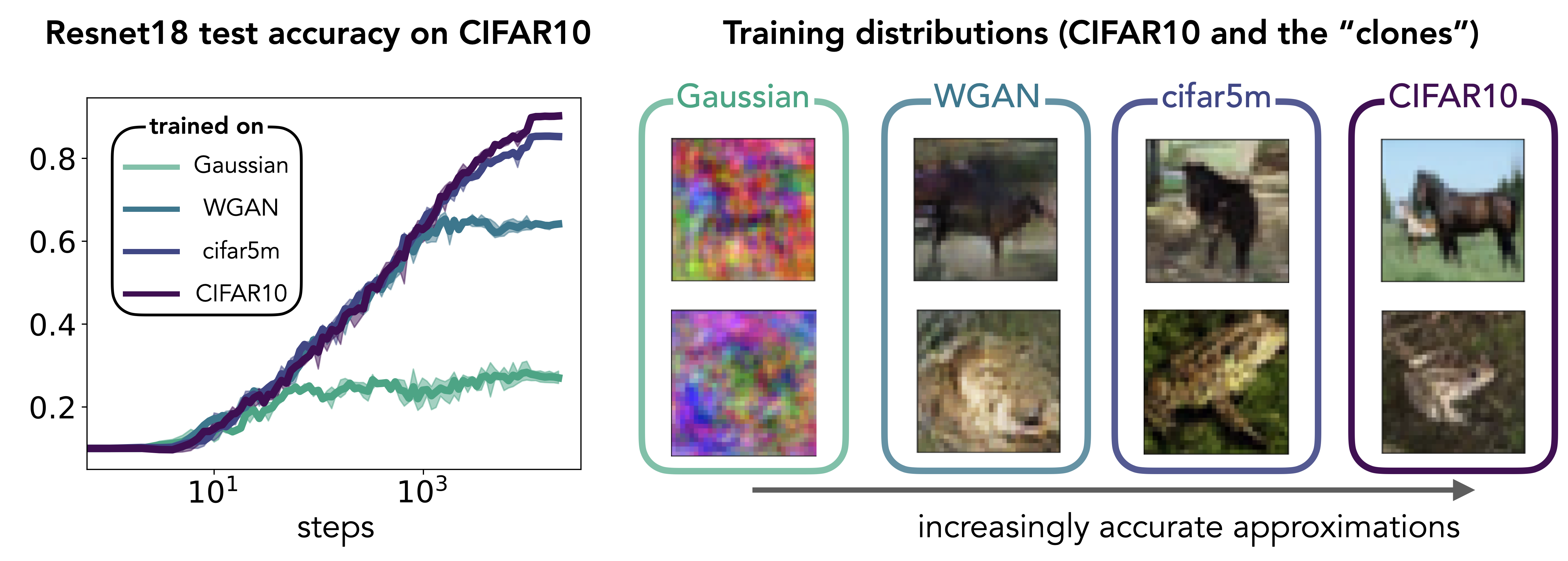}}%
  \caption{\label{fig:figure1} \textbf{Distributional simplicity bias in neural
      networks.}  Test accuracy of a ResNet18 evaluated on CIFAR10 during
    training with SGD on four different training data sets: the standard CIFAR10
    training set (dark blue), and three different ``clones'' of the training
    set. The images of the clones were drawn from a Gaussian mixture fitted to
    CIFAR10, a mixture of Wasserstein GAN (WGAN)~\citep{arjovsky2017wasserstein}
    fitted to CIFAR10, and the cifar5m data set of
    \citet{nakkiran2020bootstrap}. The clones form a hierarchy of approximations
    to CIFAR10: while the Gaussian mixture captures only the first two moments
    of the inputs of each class correctly, the images in the WGAN and cifar5m
    data sets yield increasingly realistic images by capturing higher-order
    statistics. The ResNet18 trained on the Gaussian mixture has the same test
    accuracy on CIFAR10 as the baseline model, trained directly on CIFAR10, for
    the first 50 steps of SGD; the ResNet18 trained on cifar5m has the same
    error as the baseline model for about 2000 steps. This result suggests that
    the network trained on CIFAR10 discriminates the images using increasingly
    higher-order statistics during training (experimental details in
    \cref{sec:cifar10-experiments}).}
\end{figure*}

We illustrate this idea in \cref{fig:figure1}, where we show the test accuracy
of a ResNet18~\citep{he2016deep} during training on CIFAR10 
(dark blue line; details in \cref{sec:cifar10-experiments}). To understand which
features of the data influence the ResNet, we trained the same network, starting
from the same initial conditions, on a training set that was sampled from a
mixture of Gaussians. Each Gaussian was fitted to one class of CIFAR10 and thus
accurately captured the mean and covariance of the images in that class, but all
the higher-order statistical information was lost. We show two example
``images'' sampled from the Gaussian mixture in \cref{fig:figure1}. We found
that the test accuracy of the ResNet \emph{trained} on the Gaussian mixture and
\emph{evaluated} on CIFAR10, (GM/CIFAR10 for short, green line) was the same as
the accuracy of the ResNet trained directly on CIFAR10 for the first
$\approx 50$ steps of SGD. In other words, the generalisation dynamics of the
ResNet is governed by an effective distribution over the training data, which is
well-approximated by a Gaussian mixture during the first 50 steps of SGD.

We also trained the same ResNet on two other approximations, or ``clones'', of
the CIFAR10 training set. The images in the WGAN clone were sampled from a
mixture of ten Wasserstein GAN~\citep{arjovsky2017wasserstein}, one for each
class, while the images of the cifar5m clone provided by
\citet{nakkiran2020bootstrap} were sampled from a large diffusion model and
labelled using a pre-trained classifier (details in 
\cref{sec:cifar10-experiments}). The three clones~--~GM, WGAN, and
cifar5m~--~constitute a hierarchy of increasingly accurate approximations of
CIFAR10: while the Gaussians only capture the mean and covariance of each input
class, the WGAN and cifar5m clones also capture higher-order statistics.
CIFAR5M is a more accurate approximation than WGAN, as can be seen from the
sharper example images in \cref{fig:figure1} and from the final CIFAR10 accuracy
of the ResNet trained on the two data sets, which is higher for cifar5m than for
WGAN.

The key result of this experiment is that the CIFAR10 test accuracies of the
ResNets during training on the different clones collapse: GM/CIFAR10 and
CIFAR10/CIFAR10, the base model, achieve the same test accuracy for about 50
steps of SGD; WGAN/CIFAR10 matches the test accuracy of the base model for about
1000 steps, and cifar5m/CIFAR10 matches the base model for about 2000 steps. The
effective training data distribution that governs the ResNet's generalisation
dynamics is therefore well-approximated by the GM, WGAN, and cifar5m datasets,
for increasing amounts of time.  We capture the essence of this experiment in
the following conjecture:
\begin{conjecture}[Distributional simplicity bias (DSB)]%
  \label{conjecture}
  A~parametric model 
  trained on a classification task using
  SGD 
  discriminates its inputs using increasingly higher-order input statistics as
  training progresses.
\end{conjecture}


In the remainder of this paper, we first demonstrate DSB in a solvable model of
a single neuron trained on a synthetic data
set 
(\cref{sec:perceptron}).  We then demonstrate DSB in a variety of deep neural
networks trained on CIFAR10, either from scratch or after pre-training on
ImageNet (\cref{sec:cifar10-experiments}) Finally, we place DSB in the context
of other simplicity biases, and highlight its implications for the principle of
Gaussian universality in learning (\cref{sec:discussion}).

\subsection{Further related work}

\paragraph{More simplicity biases in neural networks} \citet{arpit2017closer}
found empirically that two-layer MLP tend to fit the same patterns during the
first epoch of training, and conjectured that such ``easy examples'' are learnt
first by the network. \citet{mangalam2019do} later found that deep networks
trained on CIFAR10/100 first learn examples that can also be correctly
classified by shallow models, and only then start to fit ``harder'' examples.
\citet{valle-perez2018deep} showed that the large majority of neural networks
implementing Boolean functions with \emph{random} weights have low descriptional
complexity. 
\citet{achille2019critical} highlighted the importance of the initial period of
learning in models of biological and artificial learning by showing that certain
deficits in the stimuli early during training, especially those that concern
``low-level statistics'', cannot be reversed later during
learning. \citet{doimo2020hierarchical} analysed ResNet152 trained on ImageNet
and found that the distribution of neural representations across the layers
clusters in a hierarchical fashion that mirrors the semantic hierarchy of the
concepts, first separating broader -- and hence, easier? -- classes, before
separating inputs between more fine-grained classes. However, neither
\citet{valle-perez2018deep} nor \citet{doimo2020hierarchical} considered the
learning dynamics.  

Complementary to simplicity biases, the \textbf{implicit bias of
  SGD}~\citep{Neyshabur2015} describes the mechanism by which a neural network
trained using SGD ``selects'' one of the potentially many global minima in its
loss landscape.  In linear neural networks / matrix factorisation, this bias
amounts to minimisation of a certain norm of the weights at the end of
training~\citep{Brutzkus2018, Soudry2018, Gunasekar2017,
  gunasekar2018implicit,li2018algorithmic, ji2019implicit, Arora2019}. These
ideas have since been extended to non-linear two-layer neural
networks~\citep{lyu2020gradient, ji2020directional, jin2020implicit}, see
\citet{vardi2022implicit} for a recent review.  Here, we focus on the parts of
the data that impact learning throughout training, rather than on the final
weights.

\textbf{Code availability} \quad %
Code to reproduce our experiments and to include CIFAR10 clones in your
experiments can be found on GitHub
\href{https://github.com/sgoldt/dist_inc_comp}{https://github.com/sgoldt/dist\_inc\_comp}.

\section{A toy model}%
\label{sec:perceptron}

\begin{figure*}[t!]
  \centering \includegraphics[width=\linewidth]{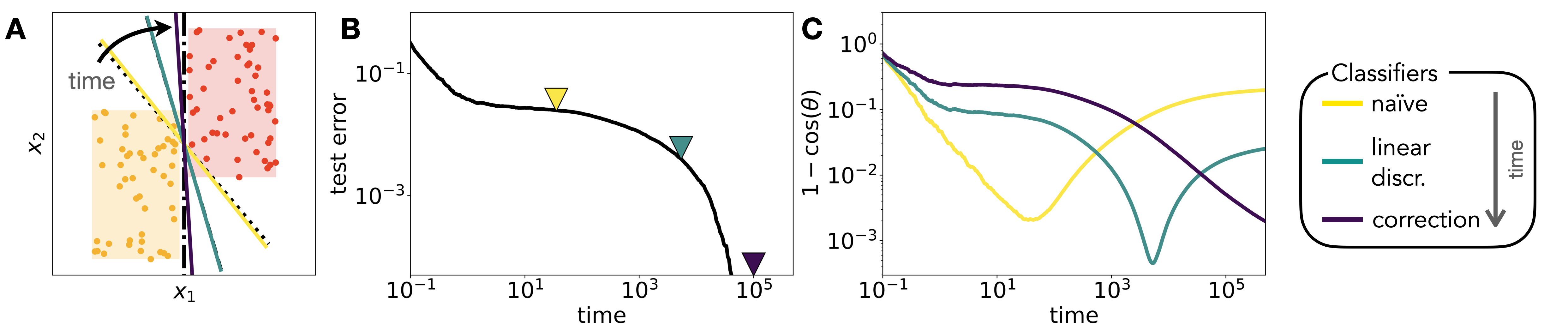}
  \vspace*{-1em}
  \caption{\label{fig:perceptron_classifier} \textbf{The decision boundary
      learnt by a perceptron takes increasingly higher-order statistics of the
      data into account during training} \quad \textbf{A:} Inputs $x=(x_i)$ are
    split into two classes (yellow vs red) which correspond to two rectangles in
    the $x_1-x_2$ plane; input components $x_{i>2}$ are sampled from a
    standard normal distribution. The coloured lines indicate the decision
    boundaries of a perceptron~\eqref{eq:perceptron} trained on this task at
    different points during training, which are indicated with triangles in
    B. The black lines are given by the naïve classifier (\cref{eq:w_0},
    dotted); the linear discriminant (\cref{eq:linear-discriminant}, dashed),
    and the oracle, which achieves perfect generalisation (dot-dashed). \quad
    \textbf{B:} Test error of the perceptron~\eqref{eq:perceptron} during
    training on the rectangular data set. \quad \textbf{C:} Alignment of the
    weight vector of the perceptron~\eqref{eq:perceptron} during training with
    the naïve classifier, the linear discriminant, and the non-Gaussian
    correction \cref{eq:correction}. Smaller values indicate smaller angles
    $\theta$ between the weight vector and the respective classifier, and hence
    larger alignment.}
\end{figure*}

We begin with an analysis of the simplest setup in which a classifier learns
distributions of increasing complexity: a perceptron trained on a binary
classification task.

\textbf{Data set} \quad %
We consider a simple ``rectangular data set'' where inputs
$x={(x^i)}_{i \le D}$ are split in two equally probable classes, labelled
$y=\pm 1$. The first two components~$x_1$ and $x_2$ are drawn uniformly from one
of the two rectangles shown in \cref{fig:perceptron_classifier}A depending on
their class. All other components are drawn i.i.d.\ from the standard normal
distribution. This data set is linearly separable with an optimal decision
boundary parallel to the $x_1$ axis, which we call the ``oracle''.

\textbf{Notation} \quad %
We follow the convention of \citet{mccullagh2018tensor} and
use the letter $\kappa$ for both moments and cumulants of the inputs. In
particular, we denote the moments of the full input distribution $p(x)$ as
$\kappa^i = \EE x^i$, $\kappa^{ij} = \EE x^i x^j$, while we denote  class-wise averages
will as $\kappa_\pm^i = \EE_\pm x^i$, and so on. We set
$\kappa^i = 0$. For cumulants, we separate indices by commas: for example,
$\kappa_+^{i,j} = \kappa_+^{ij} - \kappa_+^i \kappa_+^j$ is the covariance of
the inputs in the class~$y=+1$. The number of index partition gives the order of
the cumulant, e.g.\ $\kappa^{i,j,k}$ is the third-order cumulant of the
inputs. We summarise some useful facts on moments and
cumulants in \cref{app:tensor-methods}.

\textbf{Network} \quad %
We learn this task using a single neuron, or perceptron, whose output is given
by
\begin{equation}
  \label{eq:perceptron}
  \hat y = \sigma\left(\lambda\right), \qquad \lambda \equiv w_i x^i / \sqrt D,
\end{equation}
with weight vector $w=(w_i)\in\reals^D$ and activation function
$\sigma: \reals \to \reals$. For concreteness, we consider the sigmoidal
activation function in this section, although our argument is easily extended to
other activations. We use superscript indices for inputs and lowerscript
indices for weights, and imply a summation over any index repeated once as a
superscript and once as a subscript. We use the square loss
$\ell(\lambda, y) = {(\sigma(\lambda) - y)}^2$ for training.

\subsection{The stages of learning in the perceptron}%
\label{sec:perceptron_classifiers}

We trained the perceptron on the rectangular data set with online learning,
where we draw a new sample from the data distribution at each step of SGD,
starting from Gaussian initial weights~$w_i$ with variance~1. We show the test
accuracy of the perceptron 
in \cref{fig:perceptron_classifier}B. For the three points in time indicated by
the coloured triangles, we plot the decision boundary of the perceptron at that
time in \cref{fig:perceptron_classifier}A (the decision boundary is the line at
which the sign the perceptron output changes). Lighter colours correspond to
earlier times, so a perceptron starting from random initial weights approaches
the oracle from the left. In this section, we show that this series of
classifiers takes increasingly higher-order statistics of the data set into
account as training progresses.

We can gain analytical insight into the perceptron's dynamics by studying the
gradient flow\footnote{The index notation highlights the fact that gradient
  flow, and hence SGD, are geometrically inconsistent: they equate a
  ``covariant'' tensor (the weight) with a ``contravariant'' tensor (the
  gradient). This issue can be remedied using second-order methods such as
  natural gradient descent~\citep{amari1998natural}. Here, we focus on standard
  SGD due to its practical relevance.}
\begin{equation}
    \label{eq:gf}
    \dot{w}_i = \eta \EE \left(\sigma(\lambda) - y\right) \sigma'(\lambda) x^i,
\end{equation}
where $\EE$ indicates an average over the data distribution and we fix the
learning rate $\eta>0$. Upon expansion of the activation function around
$\lambda=0$ as $\sigma(\lambda) = \sum_{k=0}^\infty \beta_k
\lambda^k$, 
this becomes
\begin{equation}
  \label{eq:gf_expanded}
    \tau \dot{w}_i = \sum_{k=0}^\infty \EE \lambda^k x^i (\gamma_k - \tilde \beta_{k+1} y),
\end{equation}
where $\tau = 1 / \eta$, while 
$\tilde \beta_k$ and $\gamma_k$ are constants related to the Taylor expansion of
$\sigma(\lambda)$ (see \cref{app:perceptron-details}). The gradient flow updates
of the weight have thus two contributions: the first, proportional to $\gamma$,
depends only on the inputs, while the second, proportional to $\tilde \beta$,
depends on the product of inputs and their label. The interplay of this
unsupervised and supervised terms is crucial during
learning. 

\subsubsection{The first two moments of the inputs determine the initial learning dynamics}%
\label{sec:gaussian-dynamics}

We start by considering gradient flow at \textbf{zeroth order} in the weights by
truncating the sum in \cref{eq:gf_expanded} after a single term, which yields the
anti-Hebbian updates $\tau \dot{w}^{(0)}_i = - \tilde \beta_1 \EE y x_i$ (recall
that $\EE x_i = 0$). These dynamics show the typical run-away behaviour of
Hebbian learning, where the norm of the weight vector grows indefinitely, but
the weight converges \emph{in direction} as
\begin{equation}
  \label{eq:w_0}
  w^{(0)}_i \propto m^i \equiv \kappa_+^i - \kappa_-^i,
\end{equation}
which is the difference between the means of each class,
$\kappa_\pm^i = \EE_\pm x^i$. The decision boundary of the ``naïve''
classifier $m^i$ is drawn with a dashed black line in
\cref{fig:perceptron_classifier}A. Indeed, the full perceptron trained using
online SGD initially converges in the direction of this naïve classifier, as can
be seen from the overlap plot in \cref{fig:perceptron_classifier}C, which peaks
at the beginning of training (smaller values in the plot indicate smaller
angles, and hence larger alignment).

At \textbf{first order}, the steady state of gradient flow is given by
\begin{equation}
  \label{eq:gf_1}
  \gamma_1 \kappa^{ij} w^{(1)}_i = \tilde \beta_1 m^i. 
\end{equation}
At first sight, it is not clear how the \emph{global} second moment of the
inputs, $\kappa^{ij}$, where we average over both classes simultaneously, can
help the classifier separate the two classes. However, we can make progress by
rewriting the second moment as~\citep[sec.~4.1]{bishop2006pattern},
\begin{equation}
 \kappa^{ij} = \kappa_{\mathrm{w}}^{i,j} /2 + \kappa_{\mathrm b}^{i,j} / 4,
\end{equation}
with the between-class second moment $\kappa_{\mathrm{b}}^{i,j} = m^i m^j$ and
the within-class second moment
\begin{equation}
  \kappa_{\mathrm{w}}^{i,j} \equiv \EEp (x^i-\kappa_+^i)(x^j-\kappa_+^j) +
  \EE_-  (x^i-\kappa_-^i)(x^j-\kappa_-^j).
\end{equation}
Substituting these expressions into \cref{eq:gf_1} and noticing that
$\kappa_{\mathrm b}^{i,j} w_j \propto m^i$, we find that at first order,
\begin{equation}
    \label{eq:linear-discriminant}
    w^\mathrm{(1)}_i \propto {(\kappa_{\mathrm{w}})}_{i,j} m^j
\end{equation}
is a solution of the gradient flow, where ${(\kappa_{\mathrm{w}})}_{i,j}$ is the
matrix inverse of $\kappa_{\mathrm{w}}$. The classifier $w^\mathrm{(1)}_i$ is
known as Fisher's linear discriminant~\citep[sec
4.1.]{bishop2006pattern}. 
It is obtained by rotating the naïve classifier $m^i$ with the inverse of the
within-class covariance. Its decision boundary is shown by the dashed black line
in \cref{fig:perceptron_classifier}A. The linear discriminant achieves a higher
accuracy by exploiting the anisotropy of the covariances of each class. The
decision boundary of the perceptron at time $\approx 8000$ is shown in the same
plot with the green line; the overlap plot in \cref{fig:perceptron_classifier}C
also confirms that the perceptron weight moves towards the linear discriminant
after initially aligning with the naïve classifier.

\begin{figure*}[t!]
  \centering
  \includegraphics[width=\linewidth]{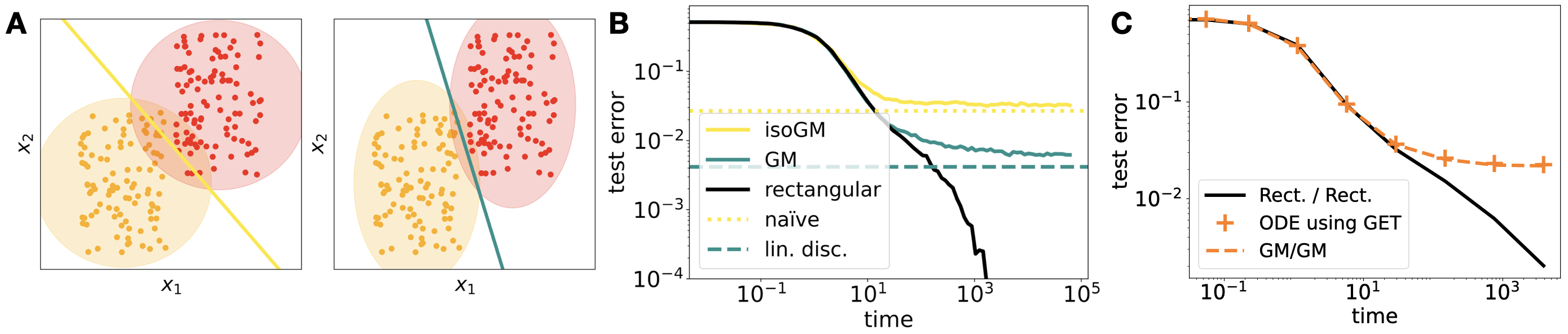}
  \vspace*{-1em}
  \caption{\label{fig:perceptron_distributions} \textbf{A perceptron learns
      distributions of increasing complexity} \quad \textbf{A}: Gaussian
    mixture approximations of the rectangular data set, using only the correct
    mean (isotropic Gaussian, left) or the correct mean and covariance
    (right). 
    \quad \textbf{B}: Test accuracy on the rectangular data set of a perceptron
    trained on the rectangular data set (black), as well as the isotropic
    (yellow) and full Gaussian mixture (green). Horizontal lines indicate the
    test accuracy of the naïve classifier, \cref{eq:w_0} and the linear
    discriminant, \cref{eq:linear-discriminant}. \quad \textbf{C}: Test error
    evaluated on the rectangular data set of a perceptron trained on the
    rectangular data set (black) and as predicted by the dynamical equations of
    \citet{refinetti2021classifying} using the Gaussian equivalence theorem
    (dashed orange line). The Gaussian theory correctly captures the dynamics of
    a perceptron trained and evaluated on the Gaussian mixture (orange
    crosses). \emph{Parameters} as in \cref{fig:perceptron_classifier}.}
\end{figure*}

\subsubsection{Higher-order correlations bias the perceptron towards the oracle}%
\label{sec:higher-order-corrections}

The linear discriminant takes the first two moments of the inputs in each class
into account, but it does not yield the optimal solution. How do higher orders
of gradient flow drive the perceptron in the direction of the oracle? We show in
\cref{app:sec:second-order} that the \textbf{second-order} term does not add
statistical information to the gradient flow. %
%
To compute the first non-Gaussian correction to the classifier, we analyse the
GF up to \textbf{third order} in the weights, which includes the fourth-order
moment of the inputs,~$\kappa^{ijkl}$.
To understand the impact of $\kappa^{ijkl}$ on the learnt classifier, we again
decompose~$\kappa^{ijkl}$ into a between-class fourth moment and a
within-class fourth moment,
\begin{equation}
  \kappa_{\mathrm w}^{ijkl} \equiv \EEp (x^i-\kappa_+^i)\cdots(x^l - \kappa_+^l)
  + \EEm (x^i-\kappa_-^i)\cdots(x^l - \kappa_-^l).
\end{equation}
We isolate the effect of the higher-order input statistics beyond the second
moment by rewriting the within-class fourth \emph{moment} as a within-class
fourth \emph{cumulant} $\kappa_{\mathrm w}^{i,j,k,l}$ and contributions from the
mean and the second moment (green):
\begin{equation}
  \kappa_{\mathrm w}^{ijkl} = \textcolor{C1}{\kappa_{\mathrm w}^{i,j,k,l}} +
  \textcolor{gaussian}{2 \kappa^{ij}\kappa^{kl}[3]}
  -\textcolor{gaussian}{\frac{1}{2} m^i
    m^j \kappa^{kl}[6]} + \textcolor{gaussian}{\frac{6}{16} m^i m^j m^k m^l}.
\end{equation}
where we use the bracket notation like $[3]$ to denote the number of possible
permutations of the indices.  We can now rewrite the steady-state of the
third-order GF as
\begin{equation}
  \label{eq:3}
  \textcolor{gaussian}{c_1 m^i} = \textcolor{gaussian}{c_2 w_j \kappa_{\mathrm w}^{ij}} + \textcolor{C1}{c_3 w_j w_k w_l
    \kappa_{\mathrm w}^{i,j,k,l}}
\end{equation}
where we separated the terms that capture the first or second moment of the
inputs (green) from the term that captures higher-order input statistics
(orange). If the data set was a mixture of
Gaussians,~$\kappa_{\mathrm w}^{i,j,k,l}$ would be zero. If we assume instead
that the inputs in each class are weakly non-Gaussian, we can solve \cref{eq:3}
perturbatively by using $c_3$ as a small parameter. To zeroth order in $c_3$,
\cref{eq:3} has the same structure as the first-order GF in \cref{eq:gf_1} and
we recover the linear discriminant, \cref{eq:linear-discriminant}. The
first-order correction is given by
\begin{equation}
  \label{eq:correction}
  w_i^{*} \propto -\textcolor{gaussian}{{(\kappa_{\mathrm{w}})}_{ij}} \textcolor{C1}{\kappa_{\mathrm{w}}^{j,k,l,m}} \textcolor{gaussian}{  w^{(1)}_k w^{(1)}_l w^{(1)}_m}
\end{equation}
and we plot the resulting classifier in violet in
\cref{fig:perceptron_classifier}A. The plot shows that the non-Gaussian
correction pushes the weight in the direction of the oracle, improving
performance. Note that if instead we computed the first-order correction with
the vanilla fourth moment~$k^{ijkl}$, the within-class fourth moment
$k_{\mathrm w}^{ijkl}$, or the vanilla fourth cumulant $\kappa^{i,j,k,l}$, we do
not obtain a correction that points in the right direction (dashed lines in
\cref{fig:correction}).

\subsection{A distribution-centric viewpoint}%
\label{sec:distribution-centric}

The perceptron did not learn an increasingly complex function during training --
its decision boundary remains a straight line throughout training. Instead, we
found that the direction of the weight vector, and hence its decision boundary,
first only depends on the means of each class, \cref{eq:w_0}, then on their mean
and covariance, \cref{eq:linear-discriminant}, and finally also on higher-order
cumulants, \cref{eq:correction}, yielding increasingly accurate predictors.

The decision boundary of the perceptron hence evolves as if the network was
trained on increasingly accurate approximations of the rectangular data set like
the ones shown in \cref{fig:perceptron_distributions}A, which accurately capture
the mean (isoGM, left), or the mean and covariance (GM, right) of each class,
resp. To illustrate this point, in \cref{fig:perceptron_distributions}D we show
the test accuracy of a perceptron \emph{evaluated on the rectangular data set},
but trained on the Gaussian clones of
\cref{fig:perceptron_distributions}A. Initially, all three perceptrons have the
same test error, which means that they use only the information in the mean to
classify samples. After the perceptron trained on isoGM converges to the naïve
classifier, the perceptrons trained on GM and the rectangular data set have the
same test error for a bit longer. The perceptron trained on the GM finally
converges to the linear discriminant, while the perceptron trained on the
rectangular data set converges to zero test error. Hence even a simple
perceptron learns distributions of increasing complexity from its data, in the
sense that it learns the optimal classifiers for increasingly accurate
approximations of the true data distribution.

\begin{figure*}[t!]
  \centering
  \includegraphics[width=\linewidth]{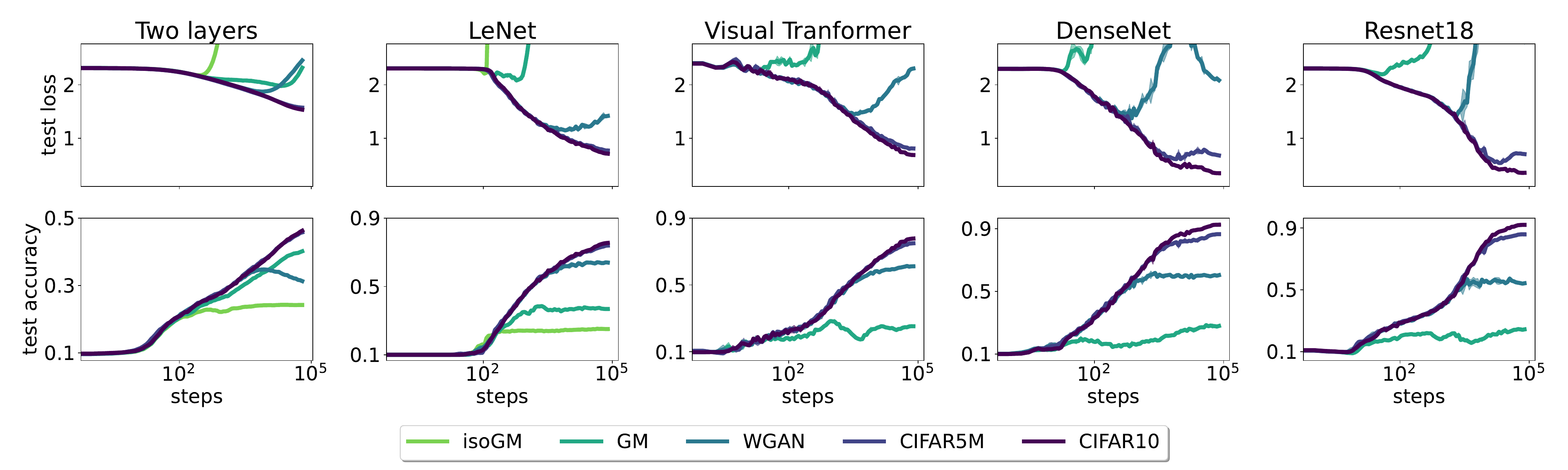}
  \vspace*{-1em}
  \caption{\label{fig:cifar10} \textbf{Neural networks trained on CIFAR10 learn
      distributions of increasingly complexity} Test loss (top) and
    classification accuracy (bottom) of different neural networks evaluated on
    CIFAR10 during training on the various approximations of the CIFAR10
    distribution described in 
    \cref{sec:cifar10-experiments}. We show the mean (solid line) over three
    runs, starting from the same initial condition each time. The shaded areas
    indicate one standard deviation over the three runs. See
    \cref{app:cifar10-architecture-training} for details on the architectures
    and training recipes.}
\end{figure*}

\subsection{The limits of Gaussian equivalence}%
\label{sec:limits-get}

The preceding analysis shows that even a simple perceptron goes beyond a
Gaussian mixture approximation of its data, provided that higher-order cumulants
of the inputs are task-relevant and improve the classifier. In other words, the
perceptron breaks the Gaussian equivalence principle, or Gaussian universality,
that stipulates that quantities like the test error of a neural network trained
on \emph{realistic} inputs can be exactly captured asymptotically by an
appropriately chosen \emph{Gaussian} model for the data. This idea allows to
characterise the performance of random features and kernel
regression~\citep{liao2018spectrum, seddik2019kernel, bordelon2020spectrum,
  mei2021generalization}, and has also been used to analyse two-layer neural
networks~\citep{goldt2020modelling, hu2022universality, loureiro2021capturing,
  goldt2022gaussian, gerace2022gaussian} and
autoencoders~\citep{refinetti2022dynamics} trained on realistic synthetic data.

\Cref{conjecture} suggests that Gaussian equivalence can always be applied to
the early period of training a neural network before the higher-order cumulants
affect the generalisation dynamics. We illustrate this phenomenon in
\cref{fig:perceptron_distributions}C, where we show the test error of a
perceptron trained on the rectangular data set as predicted by Gaussian
equivalence and the dynamical equations of \citet{refinetti2021classifying}
(orange crosses). The theory accurately predicts the test error of a perceptron
trained and evaluated on the Gaussian mixture (orange dashed line). At the
beginning of training, the theory also matches the test accuracy of a perceptron
trained and evaluated on the rectangular data (black line), but the agreement
breaks down at time $t \approx 10$. \citet{ingrosso2022data} recently made a
similar observation for two-layer networks trained on a simple model of images.


\section{Learning increasingly complex distributions on CIFAR10}%
\label{sec:cifar10-experiments}

The gradient flow analysis of the perceptron from \cref{sec:perceptron} cannot
be readily generalised even to two-layer networks. 
However, the distribution-centric point of view of
\cref{sec:distribution-centric} offers a way to test \cref{conjecture} in more
complex neural networks by constructing a hierarchy of approximations to a given
data distribution, similar to the Gaussian approximations of
\cref{fig:perceptron_distributions}.

\textbf{A hierarchy of approximations to CIFAR10} \quad %
Our goal is to construct a series of approximations to CIFAR10 which capture
increasingly higher moments. To have inputs with the correct mean per class, we
sampled a mixture with one \emph{isotropic} Gaussian for each class
(\textbf{isoGP}). Inputs sampled from a mixture of ten Gaussians (\textbf{GM}),
each one fitted to one class of CIFAR10, have the correct mean and covariance
per class, but leave all higher-order cumulants zero. Ideally, we'd like to
extend this scheme to distributions where the first $m$ cumulants are specified,
while cumulants of order greater than $m$ are zero. Unfortunately, there are no
such distributions; on a technical level, the cumulant generating function of a
distribution, \cref{eq:cumulant-generating-fn}, has either one term, two terms,
corresponding to a Gaussian, or infinitely many
terms~\citep[thm.~7.3.5]{lukacs1970characteristic}. We therefore turned to
generative neural networks for approximations beyond the Gaussian mixture.
For the \textbf{WGAN} data set, we sampled images from a mixture of ten deep
convolutional GAN~\citep{radford2015unsupervised}, each one trained on one class
of CIFAR10. We ensured that samples from each WGAN have the same mean and
covariance as CIFAR10 images; as we detail in \cref{app:synthetic-data-sets},
the key to obtaining ``consistent'' GANs was using the improved Wasserstein GAN
algorithm~\citep{arjovsky2017wasserstein, gulrajani2017improved}. We finally
also used the \textbf{cifar5m} data set provided by
\citet{nakkiran2020bootstrap}, who sampled images from the denoising diffusion
probabilistic model of \citet{ho2020denoising} and labelled them using a
Big-Transfer model~\citep{kolesnikov2020big}. 
We show one example image for each class in
\cref{fig:cifar10-clones}. 

\textbf{Methods} \quad %
We trained neural networks with different architectures on both CIFAR10 and
CIFAR10 clones like the Gaussian mixture, which we describe in detail
below. Networks were trained three times starting from the same standard initial
conditions given by \texttt{pytorch}~\cite{paszke2019pytorch}, varying only the
random data augmentations and the order in which samples were shown to the
network. Unless otherwise noted, we trained all models using vanilla SGD with
learning rate 0.005, cosine learning rate schedule~\citep{loshchilov2017sgdr},
weight decay~$5\mathrm{e}^{-4}$, momentum 0.9, mini-batch size 128, for 200
epochs (see \cref{app:cifar10-architecture-training} for
details). 

\textbf{Results} \quad %
We show the test loss and test accuracy of all the models evaluated on CIFAR10
during training on the different clones in \cref{fig:cifar10}. We found that
both a fully-connected \textbf{two-layer network} and the convolutional
\textbf{LeNet} of \citet{lecun1998gradient}, learnt distributions of increasing
complexity. Interestingly, a two-layer network trained on GM performs better on
CIFAR10 in the long run than the same network trained on WGAN. This suggests
that for a simple fully-connected network, having precise first and second
moments is more important than the higher-order cumulants generated by the
convolutions of the WGAN. The trend is already reversed in LeNet, which has two
convolutional layers. 
%
We also found a distributional simplicity bias in
\textbf{DenseNet121}~\citep{huang2017densely} and
\textbf{ResNet18}~\citep{he2016deep}, which are deep convolutional networks with
residual connections, batch-norm, etc. We also tested a \textbf{visual
  transformer} (ViT)~\citep{dosovitskiy2021image}, which is an interesting
alternative architecture since it does not use convolutions, but instead treats
the image as a sequence of patches which are processed using
self-attention~\citep{vaswani2017attention}. Despite these differences, we found
that the ViT also learns distributions of increasing complexity on CIFAR10.


\subsection{Pre-trained neural networks learn distributions of increasing
  complexity, too}%
\label{sec:pre-trained}

\begin{figure*}
  \centering
  \includegraphics[width=.9\linewidth]{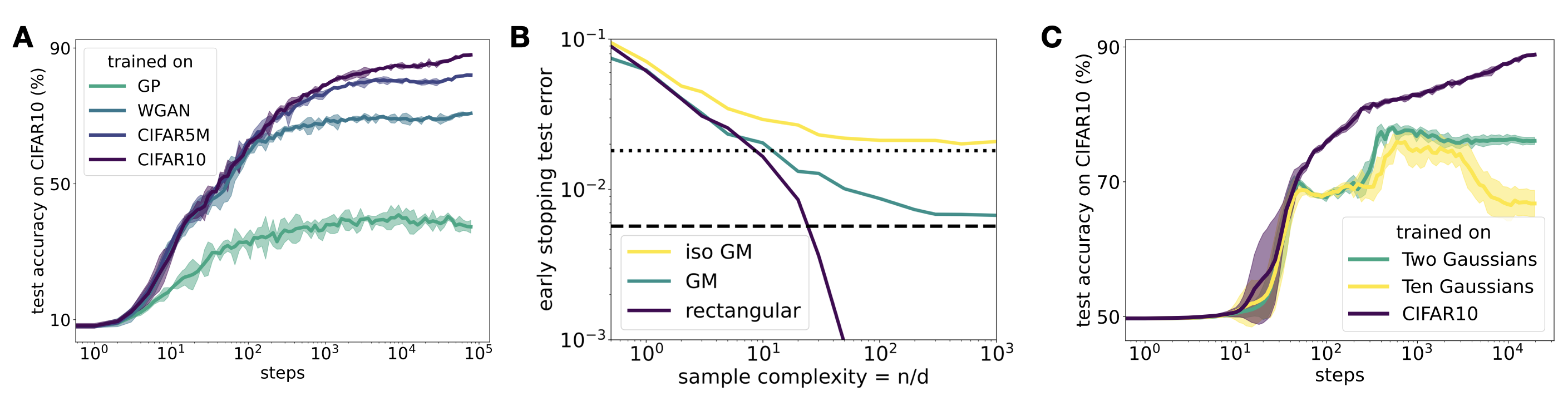}
  \caption{\label{fig:various} \textbf{A:\ Pre-trained networks learn
      distributions of increasing complexity, too.} Test accuracy of a ResNet18
    under the same protocol as in \cref{fig:cifar10}, but this time starting
    from weights that were pre-trained on ImageNet. \textbf{B:\ Distributional
      simplicity bias on finite data sets} Early-stopping test-accuracy of a
    perceptron trained using SGD on a finite number of $n$ samples, drawn from
    the rectangular data set of \cref{fig:perceptron_classifier}A (blue) and
    matching Gaussian mixtures with isotropic and anisotropic covariance (yellow
    and green, resp.). The dotted and dashed horizontal line indicate the
    performance of the naïve and linear classifier, resp. \textbf{C:\ How many
      Gaussians do you need?}  Test accuracy of LeNet evaluated on a
    coarse-grained CIFAR10 task with two classes during training on three
    different input distributions: the CIFAR10 training images, a mixture of two
    Gaussians, one for all the images in each class with the right mean and
    covariance, and a mixture of ten Gaussians, one per class with the right
    mean and covariance.}
\end{figure*}

One might argue that drawing the initial weights of the networks i.i.d.\ from
the uniform distribution biases the network towards lower-order input statistics
early during training due to the effect of the central limit theorem in
fully-connected layers. We investigated this by repeating the experiment with a
ResNet18 that was pre-trained on ImageNet, as provided by the \texttt{timm}
library~\citep{timm}. We re-trained the entire network end-to-end, following the
same protocol as above. While we found that pre-training improved the final
accuracy of the model and significantly improved the training speed, the
pre-trained ResNet also learnt distributions of increasing complexity, cf.\
\cref{fig:various}A, even though its initial weights were now large and strongly
correlated. 

This observation begs the question of what the network transferred from ImageNet
to CIFAR10, since the generalisation dynamics of the pre-trained and the
randomly initialised ResNet18 follow the same trend, albeit at different
speeds. While we leave a more detailed analysis of transfer learning using
clones to future work, we note that several papers have suggested that some of
the benefits of pre-training could be ascribed to finding a good initialisation,
rather than the transfer of ``concepts'', both in computer
vision~\citep{maennel2020what, kataoka2020pretraining} and in natural language
processing~\citep{krishna2021does}.


\section{Discussion}%
\label{sec:discussion}


The distributional simplicity bias is related to the idea of \textbf{learning
  functions of increasing complexity}, but one doesn't imply the other
directly. Any parametric model $f(x; \theta)$ that admits a Taylor-expansion
around its initial parameters~$\theta_0$ is linear in its parameters~$\theta$ at
the beginning of training, then
quadratic,~etc. 
We saw explicitly for the perceptron how this increasing degree of the effective
model that is being optimised makes the model first susceptible to class-wise
means, then class-wise covariances,~etc. However, this expansion of
$f(x; \theta)$ breaks down early during training for neural networks in the
feature learning
regime~\citep{chizat2019lazy}. 
Yet we see that the performance of cifar5m/CIFAR10 and CIFAR10/CIFAR10 agree for
almost the entire training. 

A second important distinction between DSB and the Taylor-expansion point of
view is the way in which networks are influenced by higher-order cumulants. Once
the effective model that is being optimised has degree three or higher, it
should behave differently on two data sets that have different third-order
cumulants. While we ensured that the class-wise covariances were close to each
other across the different data sets, it is reasonable to assume that already
the third- or fourth-order cumulants of WGAN or cifar5m images are not exactly
equal to each other or corresponding CIFAR10 cumulants. If the generalisation
dynamics of the neural network depended on the full third-order cumulants of
each class, the test accuracy of WGAN/CIFAR10 and cifar5m/CIFAR10 should diverge
quickly after they separate from the test accuracy of GM/CIFAR10. Instead, the
curves stay together for much longer, suggesting that only a part of the
cumulant, for example its low-rank decomposition~\cite{kolda2009tensor}, is
relevant for the generalisation dynamics at that stage. Investigating in more
detail how higher-order cumulants affect the generalisation dynamics of neural
networks is an important direction for further work.

\textbf{Learning increasingly complex distributions from finite data sets}
\quad %
Varying the data set size offers a complementary view on the
distributional simplicity bias. We explored this by training a
perceptron~\eqref{eq:perceptron} on a fixed data set with $n$ samples drawn from
the rectangular data set of \cref{sec:perceptron} using online SGD. The
early-stopping test accuracy plotted in \cref{fig:various}B shows that the
perceptron also learns distributions of increasing complexity \emph{as a
  function of sample complexity} -- for a small number of samples, the
perceptron will achieve the same test error on the rectangular data set after
training on the rectangular data (violet) or on the Gaussian mixture
approximation (green). We also note that the power-law exponent of the test
accuracy as a function of sample complexity changes roughly when the Gaussian
approximation breaks down. A theoretical treatment of this effect would requires
advanced tools such as dynamical mean-field theory~\citep{mignacco2020role,
  mignacco2020dynamical}, which leave as an intriguing avenue for further work.

\textbf{How many distributions do you need in your mixture?} \quad %
We modelled inputs using mixtures with one distribution per class, following the
structure of the gradient flow equations. We could obtain a more detailed model
of the inputs by modelling each class using a mixture of distributions. We
compared the two approaches for the two-layer network and the LeNet, which both
achieve good performance on GM/CIFAR10. We trained the networks on a
coarse-grained version of CIFAR10 (CIFAR10c): (cat, deer, dog, frog, horse) vs.\
(plane, car, bird, ship, truck)\footnote{There are six classes with living vs.\
  four classes with inanimate objects in CIFAR10, so we put birds and ships
  together because images in both classes typically have a blue background.}.
We also trained the same network on two different clones of CIFAR10c: a mixture
of two Gaussians, one for each superclass, and a mixture of ten Gaussians, one
for each input class of CIFAR10, with binary labels according to CIFAR10c. We
trained a LeNet on CIFAR10c and the two clones and found that LeNets trained on
either clone had the same test accuracy on CIFAR10c as the network trained on
CIFAR10c for the same number of SGD steps, cf.\ \cref{fig:various}C. This
suggests that the network trained on CIFAR10 started to look at higher-order
cumulants at that point, rather than at a mixture with more
distributions. However, asymptotically, the more detailed input distribution
with ten Gaussians yields better results. This observation is in line with the
results of \citet{loureiro2021learning}, who compared two-Gaussian to
ten-Gaussian approximations for a binary classification task on MNIST and
FashionMNIST and found that the ten-Gaussian approximation gave a more precise
prediction of the asymptotic error of batch gradient descent on a perceptron
(cf.\ their fig.\ 10).

\section{Concluding perspectives}%
\label{sec:conclusion}

We have found a distributional simplicity bias in an analytically solvable toy
model of a neural network, and in a range of deep networks trained on
CIFAR10. Our focus was on evaluating their test accuracy, which is their most
important characteristic in practice. To gain a better understanding of DSB, it
will be key to analyse the networks using more fine-grained measures of
generalisation, for example distributional
generalisation~\citep{nakkiran2020distributional}, and to look at the
information processing along the layers of the network. Different parts of the
networks are influenced by different statistics of their inputs; this point was
made recently by \citet{fischer2022decomposing}, who used field-theoretic
methods to 
show that Gaussian correlations dominate the information processing of internal
layers, while the input layers are also sensitive to higher-order
correlations. This finding motivates studying the structure of the internal
representations of inputs across layers for networks trained on the different
clones~\citep{alain2017understanding, bau2017network, Raghu2017,
  ansuini2019intrinsic, doimo2020hierarchical}.

The key challenge for future work is to clarify the range of validity of
\cref{conjecture} by testing it on other data sets, like
ImageNet~\citep{imagenet_cvpr09} with its rich semantic structure of
hierarchical classes. These experiments require training generative models for
each of the 1000 classes of ImageNet, so we leave this to future work. While we
have focused on correlations at the level of pixels in each image, one could
also explore different statistical ``clones'' based on correlations between
patches of pixels, or correlations in a different feature space. Likewise, it
would be interesting to consider different data modalities, such as natural
language. In the meantime, we hope that the hierarchy of approximations to
CIFAR10 will be a useful tool for further investigations on the role of data
structure in learning with neural networks.



\section*{Acknowledgements}

We thank Diego Doimo, Kirsten Fischer, Moritz Helias, Javed Lindner, Antoine
Maillard, Claudia Merger, Marc Mézard and Sandra Nestler for valuable
discussions on various parts of this work. SG acknowledges co-funding from Next Generation EU, in the context of the National Recovery and Resilience Plan, Investment PE1 – Project FAIR “Future Artificial Intelligence Research”. This resource was co-financed by the Next Generation EU [DM 1555 del 11.10.22].


\bibliography{nn}
\bibliographystyle{icml2023}

\appendix
\onecolumn

\numberwithin{equation}{section}
\numberwithin{figure}{section}

\section{Details on the theoretical analysis}%
\label{app:theory}

In this section, we summarise some useful facts and identities about moments and
cumulants in high dimensions, before giving a detailed derivation of the
theoretical results on the perceptron discussed in
\cref{sec:perceptron_classifiers}.

\subsection{Moments and cumulants in high dimensions}%
\label{app:tensor-methods}

This section summarises some basic results and identities on moments and
cumulants of high-dimensional random variables, using the notation of
\citet{mccullagh2018tensor}. We consider $D$-dimensional random variables~$x$
with components $x^1, x^2, \ldots, x^D$. We will use the superscript notation
exclusively to denote the components of $x$, which will allow us to make heavy
use of index notation and in particular, of the Einstein summation
convention. Hence we will write quadratic or cubic forms as
\begin{equation}
  a_{ij} x^i x^j = \sum_{i,j=1}^p a_{ij} x^i x^j, \qquad a_{ijk} x^i x^j x^k =
  \sum_{i,j,k=1}^p a_{ijk} x^i x^j x^k,
\end{equation}
where $a_{ij}$ are arrays of constants. For the sake of simplicity, and without
loss of generality, we take all multiply-indexed arrays to be symmetric under
index permutation\footnote{For quadratic forms $a_{ij}X^i X^j$ for example, we
  can rewrite the coefficients $a_{ij}$ into a symmetric and an asymmetric part,
  $2a = (a + a^\top) + (a - a^\top)$, with only the symmetric part
  contributing.}

We can write the moments of $x$ as
\begin{equation}
  \label{eq:moments}
  \kappa^i = \EE x^i \qc \kappa^{ij} = \EE x^i x^j \qc \kappa^{ijk} =
  \EE x^i x^j x^k \qc \mathrm{etc.}
\end{equation}
These moments are also defined implicitly by the \textbf{moment-generating function}
\begin{equation}
  \label{eq:moment-generating-fn}
  M_X(\xi) = \EE \exp(\xi_i x^i) = 1 + \xi_i \kappa^i + \xi_i\xi_j \kappa^{ij} /
  2! + \xi_i\xi_j \xi_k \kappa^{ijk} / 3! + \ldots
\end{equation}
as the partial derivatives of $M_X(\xi)$ evaluated at $\xi = 0$. The
\textbf{cumulants} are most easily defined via the \textbf{cumulant generating
  function}
\begin{equation}
  \label{eq:cumulant-generating-fn}
  K_x(\xi) \equiv \log M_x(\xi)
\end{equation}
which can similarly be expanded to yield the cumulants:
\begin{equation}
  \label{eq:cumulant-definition}
  K_x(\xi) = \xi_i \kappa^i + \xi_i\xi_j \kappa^{i,j} / 2! + \xi_i\xi_j \xi_k \kappa^{i,j,k} / 3! + \ldots
\end{equation}
Note that we use the same letter $\kappa$ to denote both cumulants and moments,
the difference being the commas, which are considered as separators for the
cumulant indices.

We can relate cumulants and moments by expanding the $\log$ of the definition of
$K_x(\xi)$~\eqref{eq:cumulant-generating-fn} and comparing terms to the
expansion \cref{eq:cumulant-definition}. Moments can be written in terms of
cumulants as
\begin{equation}
  \label{eq:moments-from-cumulants}
  \begin{aligned}
    \ka^{ij} &= \ka^{i,j} + \ka^i \ka^j, \\
    \ka^{ijk} &= \ka^{i,j,k} + (\ka^i \ka^{j,k} + \ka^j \ka^{i,k} + \ka^k
                \ka^{i,j}) + \ka^i \ka^j \ka^k\\
             &= \ka^{i,j,k} + \ka^i \ka^{j,k}[3] + \ka^i \ka^j \ka^k,
  \end{aligned}
\end{equation}
where we introduced the bracket notation $[3]$ to denote all the
partitions of the indices $i,j,k$ into two groups: $(i,jk), (j,ik)$ and
$(k,ij)$.

Similarly, we can derive expressions for the cumulants in terms of the moments
by inverting the equations above. We find that
\begin{equation}
  \label{eq:cumulants-from-moments}
  \begin{aligned}
    \ka^{i,j} &= \ka^{ij} - \ka^i \ka^j, \\
    \ka^{i,j,k} &= \ka^{ijk} - \ka^i \ka^{jk}[3] + 2\ka^i \ka^j \ka^k.
  \end{aligned}
\end{equation}

\subsection{Detailed calculations for the perceptron}%
\label{app:perceptron-details}

We start again from the gradient flow equation,~\cref{eq:gf}. Expanding the
activation function as $\sigma(\lambda) = \sum_{k=0}^\infty \beta_k \lambda^k$
yields $\sigma'(\lambda) = \sum_{k=0}^\infty \tilde \beta_{k+1} \lambda^k$ for
the derivative with
$\tilde \beta_k = k \beta_k = k \, \sigma^{(k)}(\lambda) |_{\lambda=0} $. The
gradient flow \cref{eq:gf} up to order $K$ then becomes
\begin{equation}
  \tau \dot{w}_i = \sum_{k=0}^K \EE \lambda^k x^i (\gamma_k - \tilde \beta_{k+1} y),
\end{equation}
where $\tau = 1 / \eta$, and $\sigma^{(k)}$ denotes the $k$th derivative, while
$\gamma_k$ is a linear combination of $\beta_k$:
$\gamma_0=\beta_0 \beta_1, \gamma_1 = \beta_1^2 + 2 \beta_0 \beta_2$, etc.

\subsubsection{Zeroth order: the naïve classifier} 

For $K=0$, we have
$\tau \dot{w}^{(0)}_i = x_i \gamma_0 - \tilde \beta_1 y x_i$. Under expectation,
the first term goes to zero, and we are left with the anti-Hebbian updates
$\tau \dot{w}^{(0)}_i = - \tilde \beta_1 \EE y x_i$, which show the typical
run-away behaviour of Hebbian learning, while the weight converges in direction
to
\begin{equation}
  \label{app:eq:w_0}
  w^{(0)}_i \propto m^i \equiv \kappa_+^i - \kappa_i^i,
\end{equation}
which is the difference between the mean of each class.

\subsubsection{First order: the linear discriminant}

For $K \le 1$, we find that after averaging
\begin{align}
    \begin{split}
    \label{app:eq:perceptron-order-1}
    \tau \dot{w}_i =&  - \tilde \beta_1 (\kappa_+^i - \kappa_-^i) \\
    & + w_j (\gamma_1 \kappa^{ij} - \tilde \beta_2 (\kappa_+^{ij} - \kappa_-^{ij}))
    \end{split}
\end{align}
where $\kappa^{ij}$ is the second moment of the inputs, while $\kappa_\pm^{ij}$
are the second moments for each class (cf.\ \cref{sec:perceptron}). Note that in
the rectangular data set, both classes have the same the same covariance matrix,
$\kappa_+^{i,j} - \kappa_-^{i,j} = 0$, which makes the $\tilde \beta_2$ term
vanish. 
A pattern emerges in the updates: while the
unsupervised terms $\propto \gamma$ contain moments of the inputs averaged over
the whole distribution~$p(x)$, the supervised terms ($\propto \beta$) depend on
the class-wise differences between moments.

In the steady state, we find that
\begin{equation}
  \label{app:eq:w_2}
  \gamma_1 \kappa^{ij} w_j = \tilde \beta_1 m^i
\end{equation}
At first sight, it is not clear how the \emph{global} second-moment of the
inputs $\kappa^{ij}$, which averages over both classes, can help with
discriminating the two classes. However, we can make progress by rewriting the
total second moment of the inputs as
\begin{equation}
  \kappa^{ij} = \kappa^{i,j} =\kappa_{\mathrm{w}}^{i,j} /2 + \kappa_{\mathrm b}^{i,j} / 4,
\end{equation}
where we introduced the between-class covariance~\citep[sec.~4.1]{bishop2006pattern},
\begin{equation}
  \kappa_{\mathrm{b}}^{ij} \equiv (\kappa_+^i - \kappa_-^i) (\kappa_+^j - \kappa_-^j)
  = m^i m^j,
\end{equation}
and the within-class covariance,
\begin{align}
    \label{app:eq:S_w}
    \begin{split}
      \kappa_{\mathrm{w}}^{i,j} &\equiv \EEp (x^i-\kappa_+^i)(x^j-\kappa_+^j) +
      \EE_-  (x^i-\kappa_-^i)(x^j-\kappa_-^j) \\ 
        &= \kappa_+^{ij} - 2 \kappa_+^i  \kappa_+^j
    \end{split}
\end{align}
Substituting these expressions into \cref{app:eq:w_2} and noticing that
$\kappa_{\mathrm b}^{i,j} w_j \propto m^i$, we find that the linear discriminant
\begin{equation}
  w^\mathrm{(1)}_i \propto {(\kappa_{\mathrm{w}})}_{ij} m^j
\end{equation}
is a solution to the steady-state of gradient flow up to first order. This
classifier performs better than the naïve classifier~$w_i \propto m^i$, since
rotating the naïve classifier with the inverse of the within-class covariance
brings the classifier closer to the oracle (cf.\
\cref{fig:perceptron_classifier}).

\begin{figure*}[t!]
  \centering
  \includegraphics[width=.66\linewidth]{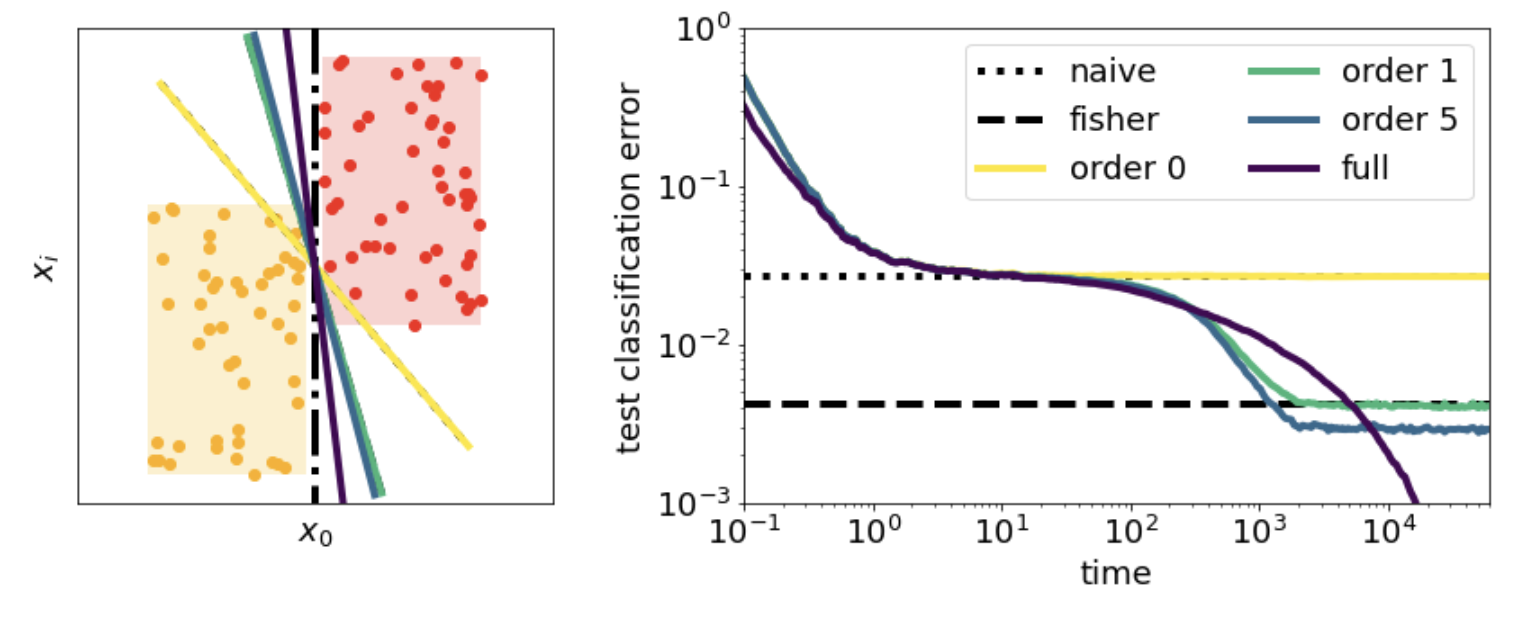}
  \caption{\label{fig:truncated_gf} \textbf{Evolution of truncated gradient
      flow}. \emph{Left}: Rectangular data set together with the weight vectors
    obtained from integrating the truncated gradient flow; colours correspond to
    the orders indicated in the legend of the right plot. \emph{Right}:
    Evolution of the test accuracy of the perceptron when the weight follows the
    truncated gradient flow at the given order. Horizontal lines indicate the
    test accuracy of the naïve classifier, \cref{eq:w_0}, and of the linear
    discriminant, \cref{eq:linear-discriminant}. Setup and parameters like in
    \cref{fig:perceptron_classifier}.}
\end{figure*}

\subsubsection{The second order does not yield new statistical information}%
\label{app:sec:second-order}

For $K \le 2$, the steady state of the gradient flow is
\begin{equation}
  \label{app:eq:perceptron-order-2}
  \tilde \beta_1 (\kappa_+^i - \kappa_-^i) =
  w_j \left(\gamma_1 \kappa^{ij} - \tilde \beta_2 (\kappa_+^{ij} - \kappa_-^{ij})\right)
  + w_j w_k \left(\gamma_2 \kappa^{ijk} - \tilde \beta_3 (\kappa_+^{ijk} - \kappa_-^{ijk}) \right)
\end{equation}
While the third-order moment $\kappa^{ijk}=0$ due to symmetry, the difference
between class-wise third-order moments $\kappa_+^{ijk} - \kappa_-^{ijk}$ is
not. We can however express this difference in terms of global quantities, i.e.\
those pertaining to the full input distribution $p(x)$, by using the identities
 \begin{equation}
  \label{eq:identities}
  \kappa_\pm^i = \pm\, m^i / 2 \qand \kappa_\pm^{ij} = \kappa^{ij}.
\end{equation}
The latter follows from the fact that by construction, the
two cumulants $\kappa_\pm^{i,j}$ are equal to each other. We find that
\begin{equation}
  \kappa_+^{ijk} - \kappa_-^{ijk} = m^i \kappa^{jk} [3] - \frac{1}{2} m^i m^j m^k,
\end{equation}
where we use the bracket notation $m^i \kappa^{jk}[3]$ to denote all the
permutations of the indices, cf.\ \cref{eq:moments-from-cumulants}. The steady
state hence becomes
\begin{equation}
  \tilde \beta_1 m^i = \gamma_1 w_j \kappa^{ij} + \tilde \beta_3 w_j w_k \left(m^i \kappa^{jk} [3] - \frac{1}{2} m^i m^j m^k\right).
\end{equation}
Upon contracting the terms proportional to $\tilde \beta_3$, we see that these terms
are all proportional either to $m^i$ or to $w_j \kappa^{ij}$. We showed above
that the linear discriminant is a solution to this equation, so going to second
order does not bring any additional statistical information that would change,
or even improve, the classifier.

\begin{figure}[t!]
  \centering
  \includegraphics[width=.5\linewidth]{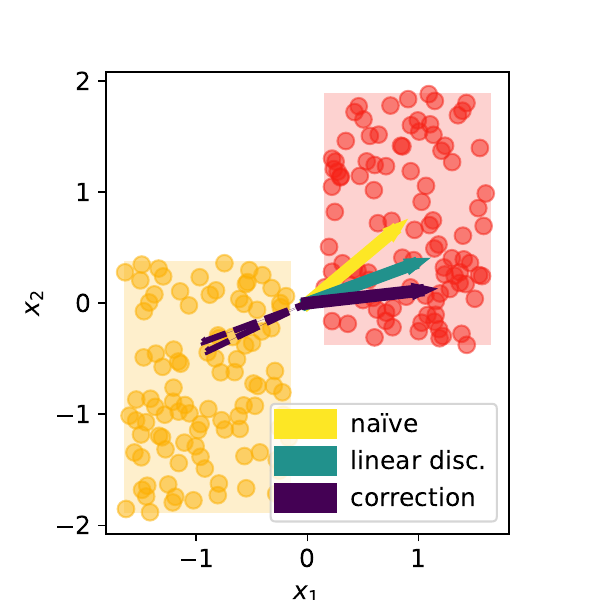}
  \caption{\label{fig:correction} \textbf{Non-Gaussian correction to the linear
      discriminant drives the perceptron towards the oracle}.  Rectangular data
    set together with the naïve classifier, \cref{app:eq:w_0}, and the linear
    discriminant, \cref{eq:linear-discriminant}. The latter only takes the
    Gaussian statistics of each class into account and does not yield the
    optimal classifier, which would be parallel to the $x_1$-axis. We show that
    the first non-Gaussian correction, \cref{app:eq:correction}, which we
    compute perturbatively in \cref{app:correction}, drives the classifier
    towards this oracle. The dashed violet lines show ``corrections'' that are
    computed with the wrong fourth-order tensor, as discussed after
    \cref{eq:correction}.}
\end{figure}

\subsubsection{Third order: the first non-Gaussian correction to the linear discriminant}%
\label{app:correction}

At \textbf{third order} in the weights however, the non-Gaussian statistics
start to impact the dynamics. The steady state is given by
\begin{equation}
  \label{app:eq:gf-s-order3}
  \tilde \beta_1 m^i = \gamma_1 w_j \kappa^{ij} + \tilde \beta_3 w_j w_k
  \left(m^i \kappa^{jk} [3] - \frac{1}{2} m^i m^j m^k\right) + \gamma_3 w_j w_k
  w_l \kappa^{ijkl}
\end{equation}
since the local terms cancel out, $\kappa_+^{ijkl} - \kappa_-^{ijkl} = 0$. Our
goal is now to understand how the fourth-order moment $\kappa^{ijkl}$ changes
the weight vector in the steady state. Following the logic of the analysis that
led us to the linear discriminant, we first decompose the fourth-order moment
into a within-class fourth-order moment $\kappa_{\mathrm w}^{ijkl}$, and
contributions from lower orders:
\begin{equation}
  \label{eq:1}
  \kappa^{ijkl} = \frac{1}{2} \kappa_{\mathrm w}^{ijkl} + \frac{1}{8}
  \kappa_{\mathrm w}^{ij}  m^k m^l[6] + \frac{1}{16} m^i m^j m^k m^l,
\end{equation}
where
\begin{equation}
  \kappa_{\mathrm w}^{ijkl} \equiv \EEp (x^i-\kappa_+^i)(x^j - \kappa_+^j)(x^k
  - \kappa_+^k)(x^l - \kappa_+^l) + \EEm (x^i-\kappa_-^i)\cdots(x^l - \kappa_-^l).
\end{equation}
After contraction with the weights, the second and third term in \cref{eq:1}
also appear in the equation that yields the linear classifier, so they do not
change the direction of the final classifier. Instead, we need to focus on the
within-class fourth moment. If the data set was a mixture of Gaussians, instead
of a mixture of rectangles, the perceptron would converge to the linear
discriminant at any order of gradient flow. So while $\kappa_{\mathrm w}^{ijkl}$
would be non-zero for the mixture of Gaussians, it would not change the
direction of the classifier. We have hence to split $\kappa_{\mathrm w}^{ijkl}$
into a Gaussian and a non-Gaussian part,
\begin{equation}
  \kappa_{\mathrm w}^{ijkl} =   \kappa_{\mathrm w}^{i,j,k,l} + \kappa_{\mathrm
    w, G}^{ijkl},
\end{equation}
where $\kappa_{\mathrm w, G}^{ijkl}$ is defined as 
\begin{align}
  \kappa_{\mathrm {w, G}}^{ijkl} &\equiv \underset{+, \mathrm G}{\EE}
  (x^i-\kappa_+^i)\cdots (x^l - \kappa_+^l) + \underset{-, \mathrm G}{\EE}
                                          (x^i-\kappa_-^i)\cdots(x^l -
                                          \kappa_-^l) \\
  &= 2 \kappa^{ij}\kappa^{kl}[3] - \frac{1}{2} m^i m^j \kappa^{kl}[6] +
      \frac{6}{16} m^i m^j m^k m^l
\end{align}
and $\underset{\pm, G}{\EE}$ denotes the average over a Gaussian distribution
with the same mean and covariance as the rectangular distribution corresponding
to $y=\pm1$.

Inserting these expressions into the steady-state gradient flow
equation~\cref{app:eq:gf-s-order3}, we find
\begin{equation}
  \label{app:eq:gf-s-order3-contracted}
  c_1(w) m^i = c_2(w) w_j \kappa_{\mathrm w}^{ij} + c_3(w) w_j w_k w_l
  \kappa_{\mathrm w}^{i,j,k,l},
\end{equation}
where the constants $c_1, c_2, c_3$ depend on the weight through the
contractions. For a fixed $c_1, c_2, c_3$, we can solve the equation. To
determine the direction of the weight, this is enough; for an exact solution,
one would need to substitute the solution for~$w$ with fixed constants back into
\cref{app:eq:gf-s-order3-contracted} and find a self-consistent equation for the
constants.

Here, we are only interested in understanding the direction of the weight
vector, so we simply fix the constants. We further assume that the data are only
weakly non-Gaussian, making $c_3$ a \emph{small} parameter that allows us to
solve the cubic equation \cref{app:eq:gf-s-order3-contracted} perturbatively. By
expanding the weight as
\begin{equation}
  \label{eq:12}
  w_i = w_i^{(1)} + c_3 w_i^{(2)} + \mathcal{O}(c_3^2),
\end{equation}
we have that $w^{(1)} \propto {(\kappa_{\mathrm{w}})}_{ij} m^j$ is the linear
discriminant by construction. The first-order correction is then given by
\begin{equation}
  \label{app:eq:correction}
  w_i^{(2)} \propto -{(\kappa_{\mathrm{w}})}_{ij} \kappa_{\mathrm{w}}^{j,k,l,m}
  w^{(1)}_k w^{(1)}_l w^{(1)}_m,
\end{equation}
which we plot in \cref{fig:correction} together with the naïve classifier and
the linear discriminant, $w^{(1)}$. From the plot, it becomes clear that this
first non-Gaussian correction to the classifier pushes the weight in the
direction of the oracle.

Note that if we instead compute the first-order correction with the full fourth
moment $\kappa^{ijkl}$ or even the within-class fourth moment
$\kappa_{\mathrm w}^{ijkl}$, we obtain corrections that point roughly in the
opposite direction of the linear discriminant (dashed lines in
\cref{fig:correction}). If instead we use the global fourth-order
cumulant~$\kappa^{i,j,k,l}$ to compute the correction, we obtain the linear
discriminant.

\section{Details on the CIFAR10 experiments}%
\label{app:cifar10-details}

\subsection{Synthetic data sets}%
\label{app:synthetic-data-sets}

\begin{figure*}[t!]
  \centering
  \includegraphics[width=\linewidth]{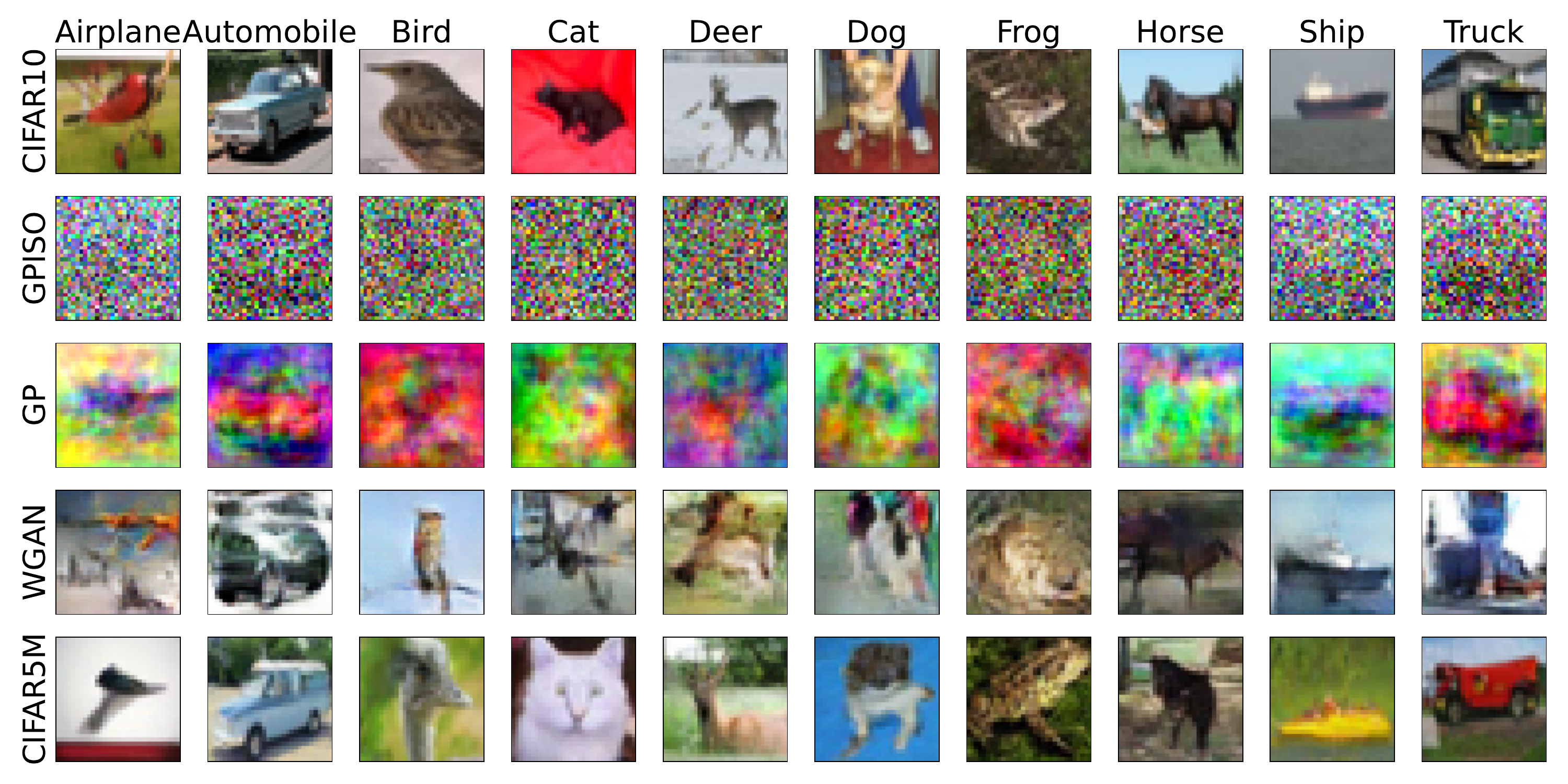}
  \caption{\label{fig:cifar10-clones} \textbf{Approximations of CIFAR10 of
      increasing complexity} We show a randomly drawn example image of each of
    the 10 classes of CIFAR10~\citep{krizhevsky2009learning} in the top
    row. Below, we show an image drawn at random from each class of four
    different approximations of CIFAR10: a mixture of one isotropic Gaussian per
    class (``isoGM''), a mixture of one Gaussians per class with the correct
    mean and covariance (``GM''), a mixture of one improved Wasserstein
    GAN~\citep{arjovsky2017wasserstein, gulrajani2017improved} per class. We
    also used the cifar5m data set of~\citet{nakkiran2020bootstrap}, who
    sampled images from a denoising diffusion model~\citep{ho2020denoising} and
    labelled it with a BigTransfer model~\citep{kolesnikov2020big}. 
  }
\end{figure*}

We constructed several approximations to the original CIFAR10 data
set~\cite{krizhevsky2009learning} for the experiments described in
\cref{sec:cifar10-experiments}. Here, we give details on how we sampled these
data sets. The code for sampling them can be found on the GitHub repository
accompanying the paper.

\paragraph{Gaussian mixtures} We constructed two Gaussian mixtures to
approximate CIFAR10. For each colour channel, we sampled from a mixture of 10
Gaussians, with one Gaussian for each class. The Gaussians had the correct mean
per class. For the isotropic Gaussian (``isoGM''), the covariance for each class
was the identity matrix times the class-wise standard deviation of the
inputs. For the correct second-order approximation (``GM''), we sampled from a
mixture of one Gaussian per class with the correct mean and covariance. We
constrained the values of the Gaussian samples to be in the range $[0,255]$ by
setting all negative values to 0 and all values larger than 255 to 255.

\begin{figure*}[t!]
  \centering
  \includegraphics[width=\linewidth]{{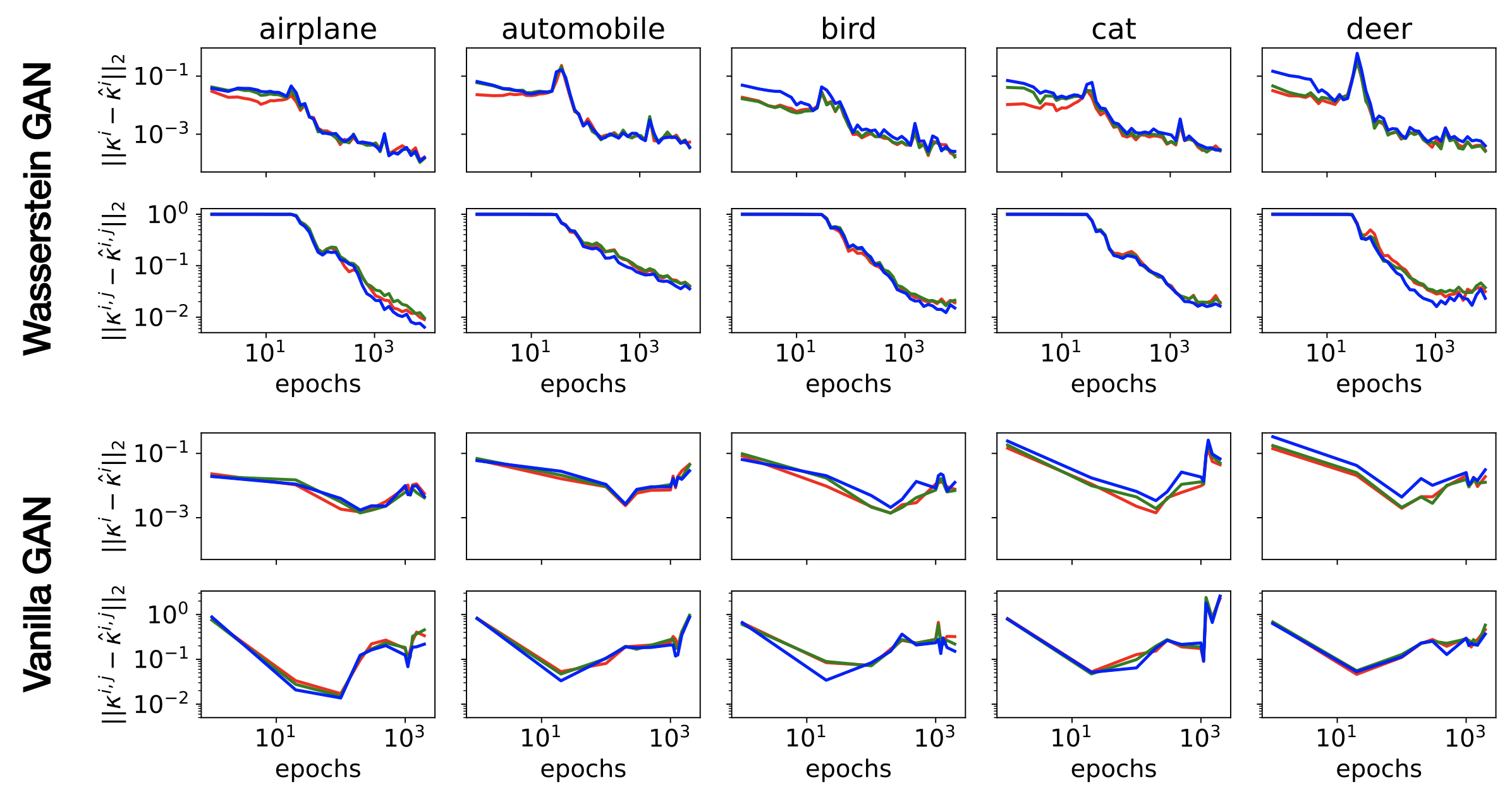}}
  \caption{\label{fig:gan-statistics} \textbf{Evolution of the mean and
      covariance of GAN-generated images during training}. We plot the
    mean-squared difference between in the mean $\kappa^i$ and covariance
    $\kappa^{i,j}$ between CIFAR10 images and samples from a GAN trained on
    images from that class, $\hat \kappa^i$, throughout training of the GAN. The
    top and bottom two rows show results for the Wasserstein and vanilla GAN,
    respectively. We only show the first five classes, results for the other
    five classes were similar. Different colours correspond to the three
    different colour channels.}
\end{figure*}

\paragraph{WGAN} We sampled images from a mixture of deep convolutional GANs,
using the architecture of \citet{radford2015unsupervised}. Each GAN was trained
on a single class of the CIFAR10 training set. Interestingly, we found that GANs
trained with the ``vanilla'' algorithm~\cite{goodfellow2014generative} yield
statistically inconsistent images: while visually appealing, the images do not
have the right covariance, and sometimes not even the right mean. We therefore
resorted to training the GANs using the improved Wasserstein GAN
algorithm~\cite{arjovsky2017wasserstein, gulrajani2017improved}. We show how the
mean and the covariance of the samples drawn from GANs trained in this way do
converge towards the mean and covariance of the CIFAR10 images, see
\cref{fig:gan-statistics}. These results are in line with an earlier theoretical
study by \citet{cho2019wasserstein} who showed that a linear Wasserstein GAN
will learn the principal components of its data. We validated the data set using
a Resnet18~\cite{he2016deep} that we trained to 94\% test accuracy on CIFAR10
following the recipe of \citet{minimal_cifar10}. This Resnet18 classified the
WGAN data set with an accuracy of 80\%.

\paragraph{cifar5m} The cifar5m data set was provided by
\citet{nakkiran2020bootstrap}. Images were sampled from the denoising diffusion
probabilistic model of \citet{ho2020denoising}, which was trained on the CIFAR10
training set. Labels were obtained from a Big-Transfer
model~\cite{kolesnikov2020big}, specifically a \texttt{BiT-M-R152x2} that was
pre-trained on ImageNet and fine-tuned on to
CIFAR10. \citet{nakkiran2020bootstrap} reported that a ResNet18 that achieved~95\% accuracy on CIFAR10/CIFAR10 achieves 89\% on cifar5m/CIFAR10, making
cifar5m a more accurate approximation to CIFAR10 than the WGAN data set by this
measure. 

\subsection{Architectures and training procedures}%
\label{app:cifar10-architecture-training}

For our experiments on CIFAR10 (\cref{sec:cifar10-experiments}), we used the
following architectures.

\begin{description}
\item[Two-layer network] Two fully-connected layers acting on greyscale images;
  ReLU activation function, 512 neurons on CIFAR10, 2048 hidden neurons on
  CIFAR100. \emph{Parameters}: Trained with SGD with learning rate 0.005, weight
  decay $5e^{-4}$, no momentum, mini-batch size 64.
\item[LeNet] LeNet5 architecture of \citet{lecun1998gradient} with five layers;
  the final layer before the linear classifier had 84 neurons for CIFAR10 and
  120 neurons for CIFAR100.
\item[DenseNet121, ViT, Resnet18] We used the implementation of these models and
  the pre-trained weights available in the \texttt{timm}
  library~\citep{timm}. For Densenet121 and Resnet18, we made a slight
  modification to the first convolutional layer, choosing a smaller kernel
  width, 3 instead of 7, at padding=1, stride=1. We also removed the first
  pooling layer before the first residual block. The idea behind this change is
  to avoid the strong downsampling of these architectures in the first
  layer. The downsampling is fine (or even necessary) for the large images of
  ImageNet, for which these networks were designed, but it removes too much
  information from the smaller CIFAR images and thus seriously degrades
  performance. We did not apply this modification for the pre-trained networks
  of \cref{sec:pre-trained}, since they were pretrained on ImageNet.
\end{description}

We trained all networks except the two-layer network using with SGD with
learning rate 0.005, cosine learning rate schedule~\citep{loshchilov2017sgdr},
weight decay $5e^{-4}$, momentum 0.9, mini-batch size 128, for 200
epochs. During training, images were randomly cropped with a padding of 4
pixels, and randomly flipped along the horizontal.

\paragraph{Comment on the accuracy improvement of ResNet18 after pre-training}
The original ResNet18 architecture achieves $88\%$ test accuracy on CIFAR10
after pre-training on ImageNet (cf.~\cref{fig:various}A). This performance
should \emph{not} be compared to the $\approx 92\%$ test accuracy of the
ResNet18 achieved in our experiments of \cref{sec:cifar10-experiments}, since we
modified the first layer of that ResNet to adjust for the image size of CIFAR10
(see above). The performance of the original ResNet18 architecture trained from
scratch without this modification when is~$85\%$.

\section{Experiments on CIFAR100}%
\label{app:cifar100}

\begin{figure*}[t!]
  \centering
  \includegraphics[width=\linewidth]{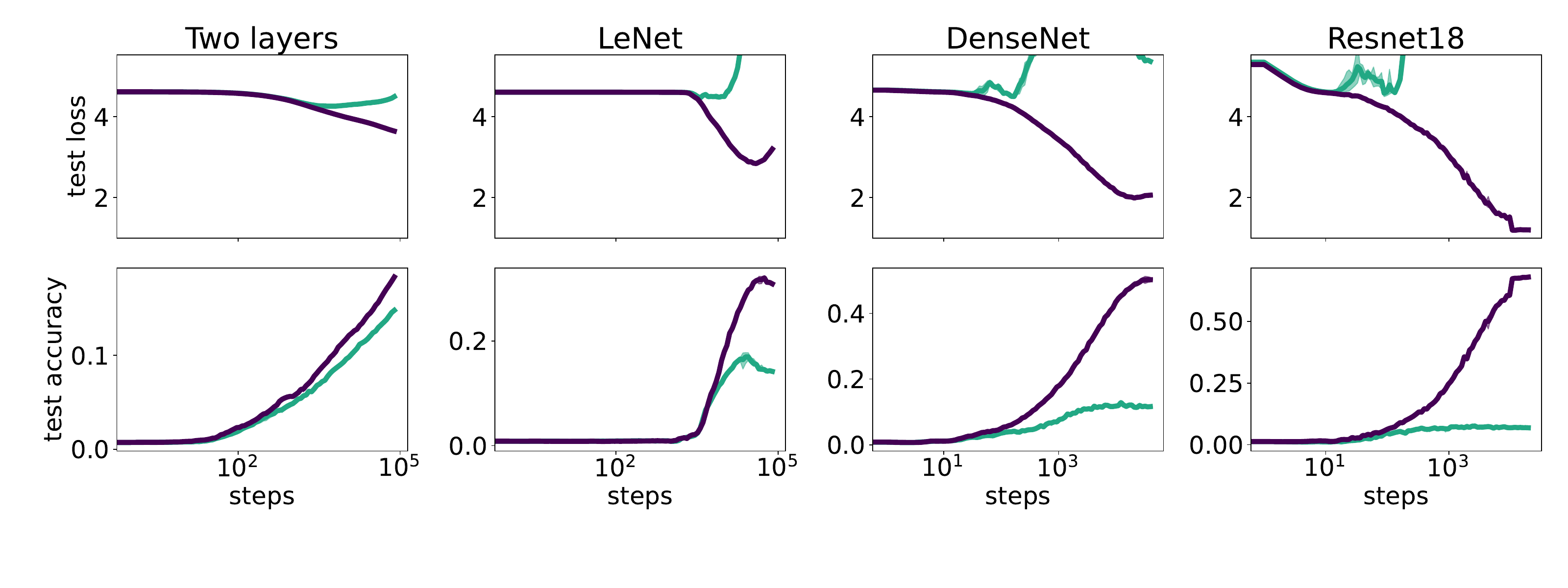}
  \caption{\label{fig:cifar100} \textbf{Test loss and accuracy of
      neural networks evaluated on CIFAR100} We repeat the experiment of
    \cref{fig:cifar10} for CIFAR100: we show the test loss and
    accuracy (top and bottom row, respectively) of various neural networks
    evaluated on CIFAR100 during training on CIFAR100 and a Gaussian clone of
    CIFAR100 with 100 Gaussians. We show the mean (solid line) over three runs,
    starting from the same initial condition each time. The shaded areas
    indicate minimum and maximum values over the three runs. Full details on the
    architectures and training recipes are given in \cref{app:cifar100}.}
\end{figure*}

We repeated the experiment on distributions of increasing complexity on
CIFAR100, which has the same image format as CIFAR10, but contains 100 classes
with 500 training and 100 test samples each~\cite{krizhevsky2009learning}. We
repeated the experimental protocol of the CIFAR10 experiments from
\cref{sec:cifar10-experiments} with the same architectures; the results are
shown in \cref{fig:cifar100} and discussed in \cref{sec:discussion}.

As we discuss in the main text, our computational facilities didn't allow to
train a separate GAN for all 100 clases of CIFAR10. We therefore only repeated
the experiments with a Gaussian clone of CIFAR100. While we see that GM/CIFAR100
yields the same test error as CIFAR100/CIFAR100 in the beginning of training for
the two-layer network and LeNet, for deeper architectures the discrepancy
appears very early during training. We leave it to further work to obtain more
accurate approximations of CIFAR100 to verify \cref{conjecture} on this data
set, too.

\end{document}